\documentclass[12pt,oneside,openany]{article}

\usepackage{amsmath,amssymb,amsfonts,stmaryrd,amsthm}
\usepackage{graphicx}
\usepackage{tabularx,booktabs}
\newcolumntype{Y}{>{\centering\arraybackslash}X}
\usepackage[center]{caption}

\title{Using Company Specific Headlines and Convolutional Neural Networks to Predict Stock Fluctuations
}

\author{Jonathan Readshaw         \and
        Stefano Giani 
}

\date{Jonathan Readshaw,
                            Department of Engineering, Durham University, Lower Mountjoy, South Road, Durham, DH1 3LE \\
              \texttt{readshawjonathan@gmail.com}   \\    
           Stefano Giani,
              Department of Engineering, Durham University, Lower Mountjoy, South Road, Durham, DH1 3LE \\
              Tel.: +44 (0) 191 33 42397\\
              \texttt{stefano.giani@durham.ac.uk}  
}

\begin{document}


\maketitle

\begin{abstract}
This work presents a Convolutional Neural Network (CNN) for the prediction of next-day stock fluctuations using company-specific news headlines. Experiments to evaluate model performance using various configurations of word-embeddings and convolutional filter widths are reported. The total number of convolutional filters used is far fewer than is common, reducing the dimensionality of the task without loss of accuracy. Furthermore, multiple hidden layers with decreasing dimensionality are employed.  A classification accuracy of 61.7\% is achieved using pre-learned embeddings, that are fine-tuned during training to represent the specific context of this task. Multiple filter widths are also implemented to detect different length phrases that are key for classification. Trading simulations are conducted using the presented classification results. Initial investments are more than tripled over a 838 day testing period using the optimal classification configuration and a simple trading strategy. Two novel methods are presented to reduce the risk of the trading simulations. Adjustment of the sigmoid class threshold and re-labelling headlines using multiple classes form the basis of these methods. A combination of these approaches is found to more than double the Average Trade Profit (ATP) achieved during baseline simulations. 
\end{abstract}

\section{Introduction}
\label{intro}
Despite suggetsions that the stock market is not predictable \cite{Malkiel_1985}, many investors and researchers seek methods that can provide market fluctuation predictions to aid investment strategy. Advances in Machine Learning (ML) and Natural Language Processing (NLP) have led to a shift in focus from technical to fundamental analysis. This new approach uses data such as news articles and historical stock prices, and is based upon the Efficient Market Hypothesis which states that an asset price reflects all available information \cite{Fama_1970}. Advances in predictive models have also led to more complex trading strategies. Most research regarding trading strategies and ML is focused on technical analysis \cite{Dash_2016,Tei_2010}, however, some works consider news and other fundamental data as part of their strategy \cite{Rachlin_2007}. The development of trading strategies based solely on fundamental data are rare throughout the relevant literature.

Early research shows no relation between headlines and stock volatility \cite{Bomfim_2003}, however the development of more advanced predictive models and availability of larger datasets has led to more accurate market trend predictions based on headlines. Although complete news articles \cite{Wuthrich_1998} or social media content \cite{Vu_2012,Bollen_2011} are used in some works, the use of headlines has become most common in this area of research due to the belief that they contain less noise than other sources of textual data \cite{Peramunetilleke_2002}. Headlines are commonly sourced from major financial news outlets such as the Wall Street Journal \cite{Werner_2004}. A wide range of prediction targets are considered throughout the relevant literature, including major indices such as the S\&P 500 \cite{Schumaker_2009} and collections of individual companies \cite{Vu_2012}. The time-span of market fluctuations analysed is also varied. For example Mittermayer \cite{Mittermayer_2004} focuses on intra-day predictions whereas long-term trends are briefly considered in the work of Ding \cite{Ding_2015}. Methods such as Support Vector Machines \cite{Schumaker_2009} and complex Decision Trees \cite{Vu_2012} remain popular for predictive tasks of this nature. These commonly use a Bag of Words (BoW) feature representation approach, where words are represented independently without consideration of word-order or context. Variations of this method include $N$-gram BoW, where phrases of length $N$ are extracted as features as opposed to single words, and Term Frequncy-Inverse Document Frequency (TF-IDF), which introduces consideration of a word's frequency within a sentence and across the entire collection. However these representations typically lead to sparsity issues when applied to a large corpus \cite{Johnson_2014}. Probabalistic approaches such as the Naive Bayes method can also be applied to tasks of this nature \cite{Ritesh_2017}. 

The development of Artificial Neural Networks (ANNs) has provided new classification and feature representation methods for text-based tasks. An ANN is a collection of nodes known as neurons that are interconnected in layers. Originally proposed by Rosenblatt \cite{Rosenblatt_1961}, the architecture is based on the transmission of signals and firing of biological neurons in a nervous system. Variations on the basic ANN architecture have been made to produce types of neural network with additional mathematical features suited to different tasks. Convolutional Neural Networks (CNNs) have gained popularity in text-based tasks. Commonly used for image recognition, CNNs  utilise a convolutional layer to detect patterns in input data that can be used for accurate classification or prediction. For example, in image detection these patterns may represent edges and shapes of a specific object depicted by its pixel values. CNNs have demonstrated state-of-the-art performance in multiple NLP tasks, including sentence classification \cite{Kim_2014} and sentence modelling \cite{Kalchbrenner_2014}. Some applications of CNNs to market prediction exist in the literature, both for major indices \cite{Ding_2015} and discrete price prediction \cite{Schumaker_2012}.

This work presents a CNN for predicting next-day stock price fluctuations of three major technology companies using headlines relating to each company. Next-day returns are used due to the inability to access the large amounts of historical intra-day stock price data required for intra-day fluctuation prediction. However the effect of news headlines has been found to resonate during the next-day period \cite{Ding_2015}. Experiments are conducted to identify an optimal model configuration for trend classification in terms of the word-embedding  and convolutional layer states. Using class predictions from these experiments, trading simulations are presented based on day-averaged predictions for each asset. Finally, modifications to both the baseline trading strategy and labelling of headlines are made with the intention of reducing risk present in simulated trading. 

\section{Model} \label{theory}
\begin{figure*}
	\centering
	\includegraphics[width=\textwidth]{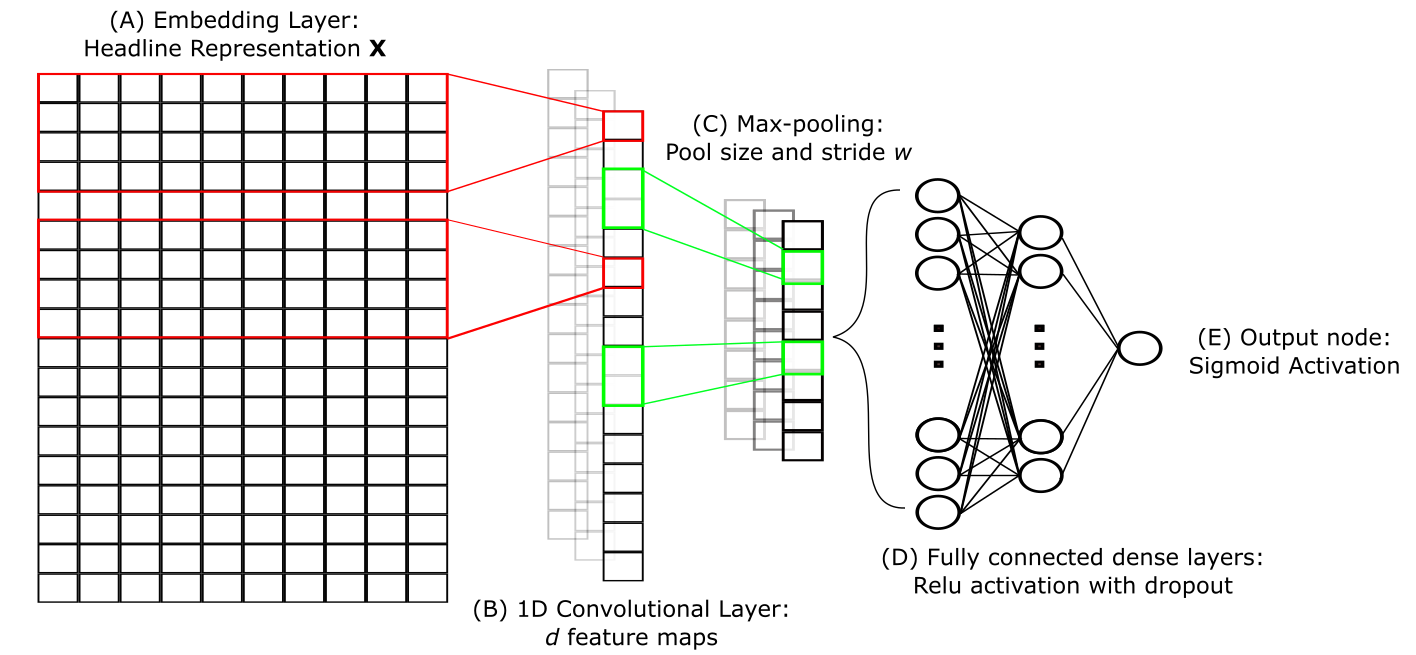}
	\caption{General model schematic where letter labelling corresponds to subsections in Section \ref{theory}. Preprocessing that precedes (A) is not included in the schematic. Note the embedding sentence representation $\textbf{X}$ is portrayed as a matrix here with dimensions $h \times p$ for ease of presentation, however in the implemented model $X \in \mathbb{R}^{hp}$ is a 1-dimensional vector. }
	\label{model}
\end{figure*}

As discussed, an ANN consists of interconnected layers of neurons. A neuron is a mathematical operator that receives one or more inputs and performs a weighted sum to generate its output. This output is often passed through an activation function that is chosen depending on both the positioning of the neuron in the network, and the task that the network is applied to \cite{Raschka_2015}. Activation functions are used to introduce non-linearity to the network, allowing for more complex mappings between inputs and outputs in the network. The network ‘learns’ by optimising each neuron's weightings to reduce the overall loss present in the system. In this work a Convolutional Neural Network (CNN) is implemented to carry out both binary and multi-class classification tasks. The overall structure of the network is comparable to those found in text-based CNN tasks throughout the relevant literature \cite{Kim_2014,Collobert_2011} but with modifications to reduce dimensionality whilst retaining accurate classification. Figure \ref{model} shows a general schematic of the network, outlining the constituent layers.

\subsection*{Preprocessing}
Raw headlines are cleaned using a tokenization algorithm that converts text to lower case, separates each sentence into its constituent words whilst removing stop-words and punctuation ('and', 'or', ':' etc.) \cite{Vijayarani_2015}. The remaining words in the collection of tokenized sentences are selected as features $u$ and form a vocabulary $V$. The selection of uni-gram features (i.e. single words) is adequate here, as phrases are evaluated using specific filter widths in the network's convolutional layer. Ordinal encoding is applied to the tokenized sentence to give vectors $\{i_1, i_2,\ldots,i_n\}$ where $i_k$ is an unique integer index corresponding to feature $u_k \in V$. Here $n \in [1,m]$ is the length of each individual tokenized sentence. To simplify implementation of the model, post-padding is applied to each vector to achieve uniform dimensionality across the set of encoded vectors. Dummy features, represented by a 0, are appended to each vector to achieve a set of $m$-dimensional vectors where $m$ is the length of the longest tokenized sentence. For each vector $\{i_1, i_2,\ldots,i_n\}$, $m-n$ zeroes are appended to the the vector. Post-padding preserves word order by only adding dummy features to the end of each vector. The result of preprocessing is a set of encoded feature vectors $\textbf{i}_{pad} \in \mathbb{R}^{m}$ that are input to the network. 

\subsection{Embedding}
Word-embeddings are utilised to represent each feature $u$ in a $p$-dimensional vector space, where $p \ll dim(V)$ to reduce the dimensionality of the problem. These representations aim to represent the semantic and syntactic context of features and the relation between similar features. Words that are interchangeable within a certain context or that often appear within close proximity of each other in a sentence are represented by similar vectors in the $p$-dimensional space and are therefore interpreted similarly by the rest of the network. Capturing context and semantics is not possible with representations such as Bag of Words or TF-IDF where features are represented with no relation to each other. In this work both randomly-initialised and pre-learned embeddings are tested. When using randomly-initiated embeddings, each $p$-dimensional vector is initiated with random values, and is tuned during the network's training epochs to form a vector representative of the feature's context in the training collection, and how it relates to other features. The random values that are used for initiation can be selected from a range of common statistical distributions. A selection of methods including random-normal and random-uniform are breifly tested during implmentation of the model. Pre-learned embeddings have been tuned on large collections of unlabelled textual data. Many well-defined collections of pre-learned embeddings formed from training on a variety of sources are available. The pre-learned embeddings used in this work are trained on a collection of 10 billion words in a Google News dataset \cite{vectors}. The embeddings are formed using the \textit{word2vec} method developed by Mikolov \cite{Mikolov_2013}. Two configurations of these pre-learned embeddings are tested in this work; \textit{static} and \textit{non-static} modes, as in the work of Kim \cite{Kim_2014}. In the \textit{static} configuration pre-learned embeddings are unaltered from their original state. However in the \textit{non-static} configuration, the network's embedding layer fine-tunes pre-learned vectors to better suit the use of a feature within the specific task. If a feature does not have a pre-learned vector representation in the \textit{word2vec} collection it is randomly initiated. Furthermore, the embedding representation of the dummy feature added during post-padding is initiated as a zero vector and remains constant throughout training for all of the configurations discussed. Each integer $i_k$ maps to a unique embedding $\textbf{x}_k \in \mathbb{R}^{p}$ representing feature $u_k \in V$. Hence for each encoded padded feature vector $i_{pad}$, the embedding layer returns the feature vector $\textbf{X} \in \mathbb{R}^{mp}$ formed by the concatenation of embeddings $\textbf{x}_1, \textbf{x}_2, \ldots, \textbf{x}_m$.

\subsection{Convolution}
In the context of textual analysis, a convolutional layer contains a number of filters that are trained to detect similar or contrasting context and sentiment in groups of adjacent words. Feature maps representing the nature of these phrases comprise of dot-product results from each sliding filter. This method can detect semantically similar phrases due to feature relations expressed using word-embeddings. The close proximity of vector representations for features in semantically similar phrases results in the calculation of similar dot-product results. Filter width $h$ dictates the length of phrases that are to be evaluated. For example, a width $h =2$ produces feature maps based on bi-grams represented in $\textbf{X}$. Consider a filter $\textbf{q} \in \mathbb{R}^{hp}$ sliding over a sentence represented by $\textbf{X}$. The first feature map element $c_1$, which represents the first phrase of $h$-adjacent words in $\textbf{X}$, is given by

\begin{equation}
c_1 = g(\textbf{q} \cdot \textbf{X}_{1, hp} + b_1)
\end{equation}

where $b_1$ is a trainable scalar bias term and $g$ is an activation function. The notation $\textbf{X}_{1, hp}$ is used to represent the elements of $\textbf{X}$ from position 1 to $hp$ inclusive. Rectified linear unit activation (relu) is used in this work, and is defined as:

\begin{equation} \label{relu}
g(x) = x^+ = \max(0,x)
\end{equation}

Each filter produces a feature map $\textbf{c} \in \mathbb{R}^{m - h - 1}$ with elements representative of each phrase in $\textbf{X}$. A single stride is implemented in this layer such that the filter slides by a single word to produce the next term in the corresponding feature map. This ensures every possible adjacent phrase of length $h$ in the headline is considered. Using a single stride, a general expression for the $k$th element in a feature map $c_k$ can be formed:

\begin{equation}
c_k = max(0, \textbf{q} \cdot \textbf{X}_{((k-1)p + 1), khp} + b_k)
\end{equation}
Using $d$ unique filters results in $d$ corresponding feature maps. Fewer filters are used in this work than is typical to reduce the dimensionality of the task and hence improve efficiency despite the presence of additional hidden layers. This work explores the effect of varying the filter width $h$ and the potential benefits of using sets of filters with different widths within a single layer. Using multiple filter widths in this layer aims to detect word patterns of different length in the original headline. Each filter is tuned during training to achieve feature maps that best represent the context of phrases and their relative importance for classification. The result of the convolutional layer is a collection of $d$ feature maps that represent groups of $h$-adjacent features in $\textbf{X}$. 

\subsection{Max-Pooling}
A max-pooling layer is used to down-sample each of the feature maps produced by the convolutional layer. Using a pool-size and stride given by $w$, each maximum value from groups of $w$ adjacent elements in a feature map is sampled. During training, filters in the convolutional layer are tuned so that feature map elements corresponding to phrases that are highly relevant for the classification of $\textbf{X}$ are large. Therefore, sampling the maximum of a group of feature elements reduces the size of the problem whilst retaining representations of phrases that are vital for classification. Consider a pool size and stride $w = 2$, the pooled feature map $\textbf{c}^{pool} \in \mathbb{R}^{\frac{1}{w}(m-h-1)}$ corresponding the the original feature map $\textbf{c}$ is given by:

\begin{equation}
\textbf{c}^{pool} = \{max(c_{1}, c_{2}),\ldots,max(c_{m-h-2}, c_{m-h-1})\}
\end{equation}

The remaining pooled feature maps are concatenated to form an input vector $\textbf{z} \in \mathbb{R}^{\frac{d}{w}(m-h-1)}$ for the hidden layers of the network. It is common for a max-over time method to be used in networks with a large number of filters $d$, where a single maximum element in each feature map is sampled \cite{Collobert_2011}. 
\subsection{Fully Connected Hidden Layers}
Two fully connected hidden layers are utilised in this work. The use of more than one fully connected layer is not common and aims to improve transmission of relationships detected by convolution through latter layers of the network. The activation of a given neuron $z_k$ based on the previous layer of  $l$ neurons $\textbf{z}_{prev} \in \mathbb{R}^{l}$ is given by 

\begin{equation} \label{activation}
z_k = \textbf{z}_{prev} \cdot \textbf{w}_k + b_k
\end{equation}

where $\textbf{w}_k \in \mathbb{R}^l$ is a trainable weights vector and $b_k$ is a trainable scalar bias term, both of which correspond to neuron $z_k$. Relu activation is applied to $z_k$ as shown by Equation \ref{relu}. Dropout is utilised in each hidden layer to reduce over-fitting and the time required to train the network. Over-fitting occurs when a model fits to the training set too closely and hence is unable to make accurate predictions based on new, unseen data. This can arise due to the high dimensional nature of text-based tasks, often referred to as the Curse of Dimensionality \cite{Friedman_1997}. Hence, a proportion of neurons at each layer are made inactive according to the specified dropout-rate to reduce training dimensionality. Each trained weights vector $\textbf{w}$ is scaled during testing to account for the probability of a neuron's exclusion due to drop-out. 

\subsection{Output Node}
For binary classification tasks undertaken in this work, a single output node is used with sigmoid activation. The sigmoid function returns the probability $\sigma (z)$ that an input headline belongs to class 1, and is defined as 

\begin{equation}
\sigma(z) = \frac{1}{1 + e^{-z}}
\end{equation}

where $z$ is the output neuron's activation as given by Equation \ref{activation}. In this work, class 1 corresponds to a next-day asset price increase, whereas class 0 represents a price decrease or continuation. Class 1 membership is allocated using a threshold $\sigma(z) \geq 0.5$. Development of a reduced risk trading strategy in this work alters this threshold to create a stricter margin. A binary cross-entropy loss function is used for binary classification tasks undertaken in this work. Loss provides a measure of how accurately the network classifies the training data using the current set of weight functions. The loss (or cost) $J(\textbf{w})$ of the current weight functions $\textbf{w}$ (including bias terms) is expressed as  

\begin{equation}
J(\textbf{w}) = -\sum\limits_k y^{(k)}log\big(\sigma(z^{(k)})\big)+\big(1-y^{(k)}\big)log\big(1-\sigma(z^{(k)})\big)
\end{equation}

where $y^{(k)}$ is the correct class label for each of the $k$ training samples. Labelling is based on the next-day stock price change of each asset, and is described in depth in Section \ref{dataset}. Weights are updated based on batches of training data through gradient-descent and backpropagation. This method updates weights and bias terms to converge towards a global cost minimum $J_{min}(\textbf{w})$. Each trainable word-embedding and convolutional filter is also tuned to better represent features and phrases using gradient descent. The global minimum represents the state in which the model is fit to the training data with the least possible error. This implies that this minimum also represents a collection of embeddings that best represent context and semantics, and a set of convolutional filters that are best at detecting word patterns vital for classification. The speed at which the model's weights approach this global minimum is determined by its learning rate. If the learning rate is too large, the model risks overshooting the global minimum, however it must be large enough to converge at a suitable rate. The model is compiled using an Adam optimiser which optimises learning rate and  gradient decay functions  \cite{Kingma_2014}. Multi-class classification is also undertaken in this work as part of trading strategy development. In these experiments the output layer of the network contains three neurons where class probability is determined by softmax activation 

\begin{equation} \label{softmax}
S(z_k) = \frac{z_k}{\sum\limits_{k=1}^3 e^{z_k}}
\end{equation}

where $S(z_k)$ is the probability of membership to the class corresponding to neuron $z_k$. The three classes used correspond to 'buy', 'inconsequential' and 'avoid' trading instructions, and are discussed further in Section \ref{sims}. A categorical cross-entropy loss function is used in multi-class tasks, and is similar in nature to the binary cross-entropy function discussed. 

\section{Dataset} \label{dataset}
This work uses two datasets from a single source to extract useful data. The first is a collection of dated headlines relating to various publicly traded from January 2007 to December 2016. Historical market data for these companies is provided in the second dataset. This market data is used to label headlines according to price change and does not contribute to any input variables for the model. It is only company-specific headlines that are used to predict market movements. From the original datasets, data relating to three technology companies; Amazon, Apple, and Microsoft, is used in this work \cite{financial_news_export}. This is due to the abundance of headlines relating to these companies, and hence they form a mock portfolio used in later trading strategy development. Along with the headline itself, the asset that the headline relates to and the date and time of its first creation are extracted from the first dataset. A relevance score is provided for each headline in the dataset where a relevance of 1 indicates the presence of the asset name in the headline. Only headlines containing the relevant asset name are used in this work, hence headlines with a relevance of 1 are extracted. 

For binary classification tasks, each headline is labelled depending on the change in next-day stock price of the relevant asset. If the asset price at market close of the next trading day is greater than the price at market open, a positive return would be made on a single day investment and hence the headline is labelled as class 1. If the asset price is unchanged or decreases over the next-day close-open period the headline is labelled as class 0. Using next-day returns has two clear benefits for evaluating model performance and its practical use. Firstly, using next-day returns allows for equal treatment of intra-hours and out-of-hours news whereas same-day returns would require substantial consideration of the time of release. Secondly, all headlines relating to an asset throughout the course of a single day can be evaluated by the model individually and subsequent next-day trading decisions can be made based on the collection of predictions for the entire day. This is explored further in the development of a trading strategy in this work.

It is not suitable to randomly split the data into training and testing sets in this context due to the fact that headlines from different sources describing the same event could appear in both sets therefore creating unwanted bias. Testing data is instead compiled of half-hourly unique headlines. The nature of the original dataset means headlines from different sources describing the same event appear within a very small time window. Hence if there is only a single headline for an asset in a half-hour period, it can be concluded that the event it describes is unique to that headline. These time-unique headlines are therefore selected for use as testing data to avoid overlapping topics in both the training and testing sets. Only testing headlines on days where each asset has at least one half-hourly unique headline are selected. This ensures that the testing date range for each asset is identical and allows for fair comparison of trading performance across the portfolio. Selecting testing data across the full range of dates in the collection aims to negate the effects of general market movements. For example if testing data was restricted to 2016 returns would be effected by global events in that year and skew the results of the various experiments undertaken in this work. Furthermore, the high-frequency nature of trades simulated  in this work ensures that long-term market trends are not important. Hence the date-range from which headlines are taken isn't an important factor when analysing the simulation results presented. There are many cases in the relevant literature of tasks similar to those undertaken in this work that simply use a random cross-validation testing split. Although these works demonstrate high accuracies, this is mainly due to a large amount of overlapping topics in training and testing creating bias. 

\section{Experimental Procedure} \label{setup}
For experiments undertaken in this work, pre-learned \textit{word2vec} embeddings of dimension $p = 300$ are used, with $d = 36$ total filters used in the convolutional layer. For configurations of the model that use filters with multiple widths, the total number of filters remains at 36. Specifically, three different widths are implemented with 12 filters each. A pool size and stride of $w = 2$ is implemented in the max-pooling layer. The number of training samples is 43060 and the number of testing samples is 7395 spanning 838 unique days. Gradients are updated based on training batches of size 32. The number of epochs and dropout-rate varies depending on the model configuration and is optimised using a grid-search \cite{Lerman_1980}.

The first collection of experiments undertaken aim to identify an optimal configuration for the CNN presented. This involves identifying an optimal filter width $h$ for the convolutional layer, comparing the effectiveness of a multiple filter width model with that of a single filter width model, and studying the effect of using randomly-initiated word-embeddings and pre-learned embeddings. The result of this experiment will be a configuration of the discussed model and associated class probabilities which are used for development of risk-minimising trading strategies. Two metrics are used to evaluate the classification performance of each model configuration; accuracy and F1-score. Model accuracy can be expressed generally as the ratio of the number of correct class predictions to the total number of predictions. F1-score is used to account for both the precision and recall of each configuration. Using a single metric to account for both of these measures allows for easier identification of an optimal configuration. Accuracy is quoted due to its common presence in relevant literature for classification tasks. Precision and recall are defined as 

\begin{equation}
PRE = \frac{TP}{FP + TP} \qquad REC = \frac{TP}{FN + TP}
\end{equation}



where $TP$ is the number of true positive predictions, $FP$ the number of false positive predictions, and $FN$ the number of false negative predictions. Further interpretation of precision is discussed in Section \ref{sims} where it forms the basis for a common metric used to evaluate day-averaged predictions as opposed to those based on individual headlines. The corresponding F1-score is defined as: 

\begin{equation}
F1 = 2 \times \frac{PRE \times REC}{PRE + REC}
\end{equation}

\section{Optimum Model Configuration} \label{optimal}
For results discussed in this section the following terms are used to refer to different configurations of the CNN outlined in Section \ref{theory}. Firstly, \textit{single-width} is used to denote a configuration that uses a single filter width $h$ in the convolutional layer, whereas \textit{multi-width} denotes a model that uses three different values of $h$. Word embedding methods are described as \textit{self-learnt} for randomly initiated vectors, and \textit{static} or \textit{non-static} for non-trainable and trainable pre-learned embeddings respectively. For example a configuration labelled as \textit{single-width self-learnt} describes a model using randomly initiated word-embeddings with a single filter width in its convolutional layer. Tables \ref{single_results} and \ref{multi_results} show the classification results, optimal filter width(s) $h$ and number of training epochs for \textit{single-width} and \textit{multi-width} implementations respectively. Although the stated accuracies are lower than in other NLP tasks such as sentence classification \cite{Kim_2014}, it is important to consider the nature of the task. Unlike other classification tasks, predicting stock price movements is heavily dependant on factors beyond the textual content of the headline such as general market movements and trader behaviour. It is common for a headline to have a positive sentiment towards a particular asset but for the price of the asset to decrease. Factors such as increased leverage and information possessed by traders that is not readily available can cause these results. However, the results obtained in this work are significantly better than random guessing (50\%) and therefore it can be stated that the various implementations of the model are able to predict short-term market trends solely based on news headlines.

\begin{table}

\begin{tabularx}{\columnwidth}{X Y Y Y}
\toprule
 & Self & Static & Non-Static\\
\midrule
Accuracy [\%] & 59.6 & 57.4 & \textbf{61.5}\\
F1-score [\%] & 57.6& 57.0& \textbf{58.7} \\
Filter width $h$ & 3& 4& 4\\
Epochs & 5& 7& 7 \\
\bottomrule
\end{tabularx}
\caption{Classification metrics, optimal filter width and number of training epochs for Single-width Implementations}
\label{single_results}
\end{table}

\subsection{Effect of Filter Width}
\begin{figure}
\centering
\includegraphics[scale=0.8]{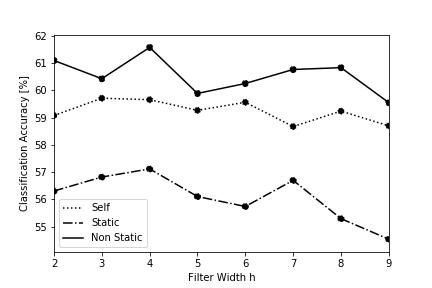}
\caption{Variation in accuracy with filter width $h$ for each \textit{single-width} model configuration.}
\label{width}
\end{figure}

\begin{figure}
\centering
\includegraphics[scale=0.8]{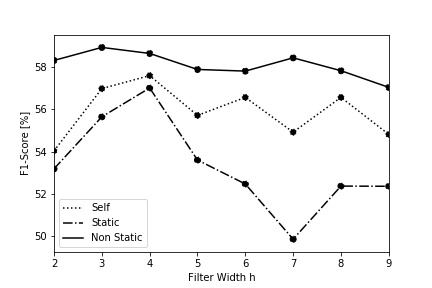}
\caption{Variation in F1-score with filter width $h$ for each \textit{single-width} model configuration.}
\label{f1_width}
\end{figure}

Table \ref{single_results} provides the optimum filter width $h$ for each \textit{single-width} implementation of the model used in this work. Using a grid-search over widths $h \in [2,9]$ the effect of this parameter on classification accuracy and F1-score is established. Figures \ref{width} and \ref{f1_width} show the variation in the relevent performance metrics with filter width $h$ for \textit{single-width} implementations of the model. As seen in Table \ref{single_results} the optimum single width is similar for each word-embedding state. The best classification results are produced by evaluating tri-gram or quad-gram phrases in the convolutional layer. These results can be explained by considering phrases within a headline that depict its sentiment. Phrases with a noun-verb-adverb structure tend to summarise the tone of a headline in its entirety. For example the phrase "\textit{shares fall sharply}" within a particular headline provides all the necessary information to make an accurate prediction regarding next-day stock price movement. Other information is commonly of little importance and can be considered noise. Hence each model demonstrates an optimum filter width of 3 or 4 where these vital phrases can be represented in their entirety. 

Each word embedding state also demonstrates good accuracy at some larger filter widths. This suggests that the network is able to represent longer phrases that provide relevant information for predicting price fluctuations. It may be the case that combinations of the discussed three and four word phrases that carry sentiment can be represented as a single feature map element. Using the results obtained, each \textit{multi-width} model implementation uses the three best feature widths in terms of F1-score for each word embedding state in the corresponding \textit{single-width} model. Selection is based on F1-score due to the relevance of precision and recall to latter trading simulations, where false positives and negatives are considered.

Table \ref{multi_results} shows the classification results for each \textit{multi-width} implementation of the model. Although the \textit{non-static multi-width} model configuration produces the best classification results for both metrics considered, the use of multiple filter widths for \textit{static} and \textit{self-learnt} embeddings results in worse classification performance than in the equivalent \textit{single-width} cases. Overall the progression to multiple filter widths produced no consistent benefit to model performance across the word-embedding states. This result can be explained by the reduced number of filters $d$ used for each optimum filter width shown in Table \ref{single_results}. As outlined in Section \ref{setup}, the number of total filters in the convolutional layer is equal for both \textit{single-width} and \textit{multi-width} implementations. For example in the \textit{single-width static} configuration there are 36 filters with width $h=3$ however in the equivalent \textit{multi-width} configuration there are only 12 filters of this width. Therefore despite the ability to represent phrases of different lengths using multiple filter widths, there are less filters available to represent the key tri-gram or quad-gram phrases discussed previously. Therefore to observe significant improvements in classification using multiple filter widths, it is expected that the number of filters for each width would have to be greater than or equal to the total number of filters used in the single-width implementation. However this increases the dimensionality of the network and time required for training. Furthermore, the ability to detect the presence of each phrase is dependant on the accurate tuning of each filter through gradient-descent.

\begin{table}

\begin{tabularx}{\columnwidth}{X Y Y Y}
\toprule
 & Self & Static & Non-Static\\
\midrule
Accuracy [\%] & 59.2 & 56.5 & \textbf{61.7}\\
F1-score [\%] & 58.3& 56.5& \textbf{59.2} \\
Filter widths $h$ & 4, 6, 8& 3, 4, 5& 3, 4, 7\\
Epochs & 5& 7& 9 \\
\bottomrule
\end{tabularx}
\caption{Classification metrics, optimal filter width and number of training epochs for multi-width Implementations}
\label{multi_results}
\end{table}

\subsection{Word-embeddings}
\begin{figure}[h]
\centering
\includegraphics[scale=0.8]{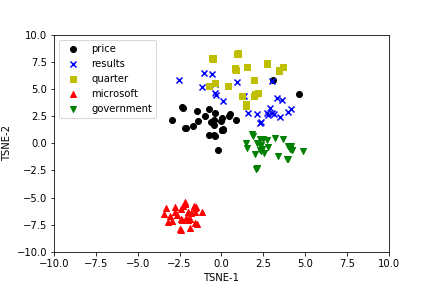}
\caption{Two-dimensional representation of the word-embeddings of the 30 most similar terms to each reference term shown in the legend. Similarity between vectors is calculated by cosine similarity. Mapping of 300-dimensional word-embeddings to two-dimensions is achieved using t-SNE.}
\label{tsne}
\end{figure}

Figure \ref{tsne} shows a projection of a sample of pre-learnt \textit{word2vec} word-embeddings on a two-dimensional space. Each point corresponds to a word  whose 300-dimensional vector representation is part of the 30 most similar vectors to the associated reference word shown in the figure legend.  In this work, vector similarity is calculated using cosine similarity.  The 300-dimensional embeddings are projected onto a two dimensional space using t-distributed stochastic neighbour embedding (t-SNE) \cite{t-SNE}. This method models each high-dimensional vector as a two-dimensional point such that similar high-dimensional vectors are modelled as similar points in the low-dimensional space. Hence the similarity of high-dimensional word-embeddings can be visualised. The clustering of points that represent vectors that are similar to a specific reference word demonstrates the principle of word-embeddings, where terms used in a common context have similar positions in the 300-dimensional word-embedding space. These terms are therefore interpreted similarly by convolutional filters. Figure \ref{tsne} also shows the formation of larger clusters containing points corresponding to different reference words. This identifies larger groups of terms that have similar context or that could appear in close proximity within a sentence. For example the overlap between clusters corresponding to 'quarter' and 'results' could arise from frequent use of the phrase 'third quarter results'. This overlapping therefore demonstrates how the use of word-embeddings not only accounts for if terms are interchangeable within a certain context but also how frequently terms appear in close proximity within a text. The isolation of terms similar to 'microsoft' (e.g. 'photoshop') is due to the specificity of these terms and the lack of any contextual relationships to the other reference words shown without fine-tuning.

For \textit{single-width} and \textit{multi-width} implementations of the model, word-embeddings in the\textit{ non-static} state lead to the best classification accuracy and F1-score. Table \ref{embeddings} provides the most similar term to the terms shown in Figure \ref{tsne} for the \textit{static} and \textit{non-static} configurations following training. It can be seen that in the \textit{non-static} state, embeddings corresponding to words that can be used in a range of contexts are tuned to better represent the context of investment news and the stock-market. For example, the similarity between 'quarter' and 'half' in the \textit{static} state is a general relation that can apply to many different contexts however the similarity between 'quarter' and 'q2' in the \textit{non-static} state suggests a relationship based on a financial context. This refinement in context allows for the model to predict market trends based on headlines with greater accuracy. Table \ref{embeddings} also shows how words typically used in a consistent context (i.e. proper nouns) undergo little refinement. This embedding refinement is demonstrated across the literature for sentence classification tasks where the target variable is solely dependent on the textual data \cite{Kim_2014}. However, it is interesting to observe similar behaviour in this work where the model is still able to refine representations based on context despite the target variable depending on factors beyond the context of the headline (e.g. overall market behaviour).

\begin{table}

\begin{tabularx}{\columnwidth}{Y Y Y}
\toprule
Term & \multicolumn{2}{c}{Most Similar Term} \\
& \textit{Static} & \textit{Non-static}\\
\midrule
'results' & 'findings' & 'earnings'\\
'price' & 'premium' & 'stock'\\
'quarter' & 'half' & 'q2'\\
'microsoft' & 'adobe' & 'adobe' \\
'government' & 'administration' & 'administration' \\
\bottomrule
\end{tabularx}
\caption{Most similar terms based on cosine similarity of word-embeddings for both \textit{STATIC} and \textit{NON-STATIC} configuration of the network's convolutional layer.}
\label{embeddings}
\end{table}

Various initiation methods for \textit{self-learnt} embeddings were tested. The best classification results were obtained using a random normal initiation, where vectors are initiated randomly with a normal distribution with mean and standard deviation equal to those of the pre-learned embeddings used in this work. The classification results using \textit{self-learnt} configurations are better than those of \textit{static} configurations for both \textit{single-width} and \textit{multi-width} models. This result suggests that in the \textit{self-learnt} configuration the model is able to suitably learn a set of embeddings that represent the context of the collection. However, \textit{self-learnt} embeddings often over-fit the context of the training data and fail to represent similarities between words that may be interchangeable in different contexts. This is because \textit{self-learnt} embeddings are formed solely on the context of the training set and cannot account for words that do not appear in training. Additionally if the context in which a word is used in a testing headline is slightly different to that found in training, the self-learnt embeddings misrepresent this due to the limited word relationships that can be established from the small training set. This results in less accurate classification than \textit{non-static} configurations where embeddings retain relations based on a variety of contexts from their initial states despite fine-tuning. For example, 'half' remains the fifth closest word to 'quarter' using \textit{non-static} embeddings. A much larger training set would be required for general context relationships to be represented in \textit{self-learnt} embeddings. These observations therefore suggest that \textit{non-static} embeddings provide the best configuration not only because of their ability to be fine-tuned to the task in question, but also because a more general context of words is retained in the embeddings allowing for better application to both unseen headlines and new tasks.

\subsection{Overall Optimal Model}
Based on the experimental results discussed in this section it can be concluded that the best configuration of the model for predicting fluctuations in next-day stock prices is a \textit{multi-width} implementation using \textit{non-static} word-embeddings. A \textit{single-width} implementation using \textit{non-static} embeddings would also be suitable for this task as the advantage of using multiple filter widths instead of a single width  is not uniform across each word-embedding state. \textit{Self-learnt} embeddings are not suitable for implementation due to their tendency to over-fit the context of training headlines and inabillty to represent words not present in training. Hence, it is expected that the embeddings would not be suitable for testing headlines relating to companies from different sectors or textual-data from a different source (e.g. Twitter). Conversely, \textit{static} embeddings are unable to suitably represent the specific context of the task. The maximum classification accuracy achieved is comparable with those achieved by state-of-the-art methods \cite{Ding_2015}, however the comparison of performance across works that use different datasets is not conclusive. Results are heavily dependant on the number and quality of headlines in the dataset used, and since there is no standard dataset that is used by the majority of works in this field it is hard to fairly compare performance. This is not the case in other fields such as image recognition, where baseline datasets exist. The next section aims to develop a trading strategy based on the \textit{multi-width non-static} model implementation to manage risk and prevent incorrect stock purchases. 

\section{Trading Simulations} \label{sims}
This section seeks to evaluate the performance of the discussed CNN model using a simple trading strategy. It is first necessary to assess whether the optimal model configuration in terms of individual headline classification results in the best trading results. The baseline strategy used in this section is as follows. For each unique day in the testing set, the mean of the sigmoid outputs $\sigma(z)$ for headlines relating to each individual asset is calculated. This mean value $\sigma(z)_{mean}$ aims to represent a prediction based on all the headlines relating to the associated asset throughout the day. If  $\sigma(z)_{mean} > 0.5$, shares in the relevant asset are purchased on the next trading day at market opening and sold at market close. Simulated returns can therefore be calculated using the next-day close-open price change of the asset. Using this strategy, a metric that evaluates the trading performance of the model can be formed based on precision. In this context, a true positive (TP) indicates a prediction $\sigma(z)_{mean} > 0.5$ where the next-day returns are positive, whereas a false positive (FP) describes such a prediction where the next-day returns are negative. Therefore, precision describes the proportion of next-day investments made based on the model's predictions that lead to a positive return. This metric is commonly referred to as the percent profitable (PP) metric in trading applications. Another common metric used is average trade profit (ATP) which evaluates the average return of executed trades. The total percentage return based on the described strategy across the entire date range present in the testing set is also used to evaluate the performance of the model. 

Table \ref{baseline} shows the trading performance of each of the model configurations discussed in Section \ref{optimal}. The \textit{multi-width} \textit{non-static} implementation of the model demonstrates the best trading performance according to each of the metrics used. Hence, it can be concluded that the optimal configuration for classification of individual headlines also provides the best trading performance for simulations considered in this work.  

\begin{table}

\begin{tabularx}{\columnwidth}{X  l c c c}
\toprule
 & & Returns [\%] & PP [\%] & ATP [\%]\\
\midrule
 & Self-learnt & 238.8	& 57.4	& 0.195\\
Single-width & Static & 221.1	&55.3	&0.185\\
 & Non-static & 272.8 & 58.5 & 0.226\\
\midrule
 & Self-learnt & 293.5	&58.0	&0.245\\
Multi-width & Static & 140.3	&53.7	&0.111\\
 & Non-static & \textbf{317.8}	&\textbf{60.9}	&\textbf{0.284}\\
\bottomrule
\end{tabularx}
\caption{Trading performance for each model configuration using the baseline trading strategy.}
\label{baseline}
\end{table}

Using these baseline results, two methods to reduce the risk in the system are proposed. In the context of this work, risk is associated with how frequently the model's predictions result in a next-day investment leading to losses, and the average loss incurred by these incorrect investments. Hence, a system with reduced risk would demonstrate higher PP and ATP metrics. It is expected that lower returns will be made using these methods due to the risk-return trade off that exists in market trading \cite{risk_return}. The methods discussed aim to provide strategies that can be used during periods of greater market volatility, where reduced risk is of greater priority than large returns. The two methods proposed involve the adjustment of the buy threshold associated with $\sigma(z)_{mean}$, and modification of the original classification task from binary to multi-class.


\subsection{Buy Threshold}
In the baseline trading strategy discussed the mean prediction for headlines relating to a specific asset throughout a single day is used to make trading decisions. The constraint $\sigma(z)_{mean} > 0.5$ is used as a baseline, where 0.5 can be considered as a buy threshold. For each individual input to the network, the closer the sigmoid output $\sigma(z)$ is to unity, the greater the certainty in the headline's membership to class 1. Therefore, increasing the buy threshold aims to select days where $\sigma(z)_{mean}$ represents greater certainty that the next-day returns will be positive. Simulations are conducted across the entire testing date range for $\sigma(z)_{mean} > t$, where $t \in [0.5, 0.9]$ is the buy threshold used.

\begin{figure}
\centering
\includegraphics[scale=0.8]{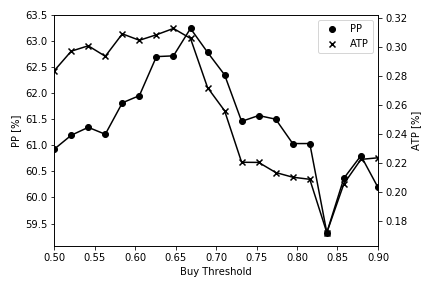}
\caption{Variation in PP and ATP with buy threshold across the entire date testing date range using a strategy developed from binary label classification.}
\label{pp}
\end{figure}

Figure \ref{pp} shows the variation in PP and ATP with buy threshold. The highest percentage of profitable trades is achieved using a buy threshold $t = 0.67$ with an improvement of 2.3\% over the baseline strategy. Despite this increase in PP the overall returns using this method are reduced. Although a higher ATP than baseline is demonstrated at some buy thresholds, the stricter margin results in less investments being made. For example using $t = 0.67$, 38.1\% less investments are made across the testing period than in the baseline case. This result demonstrates the risk-return trade-off discussed. It is expected that the strictest buy thresholds would yield the most improvement in PP and ATP, however both metrics demonstrate a decrease in these metrics for high values of $t$. This result is due to the model making incorrect predictions despite high values of $\sigma(z)_{mean}$. For example, the largest single-day loss from an investment is 11.3\%, however the model predicts $\sigma(z)_{mean} = 0.93$ based on the previous day's headlines. The effect of these incorrect predicitions with high $\sigma(z)_{mean}$, coupled with a significantly reduced number of trades, leads to lower performance metrics than in the baseline case at  values of $t > 0.75$. These results demonstrate the shortcomings of making predictions based solely on company headlines, as it is possible for the network to make a positive prediction with high certainty based on a collection of headlines but for a significant loss to be made. This is due to market activity and trends that cannot be depicted within a single headline. Therefore, to see continued risk reduction at higher buy thresholds, general economic headlines or historical price trends must also be considered by the model. Despite this, reduced risk is demonstrated across a range of moderate buy thresholds using this method. The optimum value of $t$ achieved here is similar to that shown in the work of Ding et al \cite{Ding_2015} who consider a similar method. However, their method is only applied to individual headline classification and does not consider trading decisions based on headlines from an entire day as presented in this work. 

\subsection{Modification to Multi-class Labelling}
The second approach adopted to reduce risk involves the alteration of the original classification task from binary to multi-class. In the multi-class experiments conducted, individual headlines belong to one of three classes; 'avoid', 'inconsequential' or 'buy'. Headlines relating to an asset who's stock price falls by more than 0.5\% during the next-day's trading are labelled as 'avoid'. If the next-day returns of an asset is greater than 0.5\% then corresponding headlines are labelled as 'buy'. Headlines that fall between these constraints are labelled as 'inconsequential'. This labelling system aims to encourage the model to identify investment opportunities that are likely to provide significant return. The 'inconsequential' class contains some investment opportunities with positive return, however the risk associated with them is larger than the potential reward. Whereas the mean of each sigmoid output on a single day per asset was used to make trading decisions in both the baseline and buy threshold experiments, a more complex decision process is required here. Using the softmax activation given in Equation \ref{softmax}, three probabilities, $S(z_1), S(z_2)$ and $S(z_3)$, are output by the network for each headline, corresponding to the probability of membership to each of the three classes. Here classes 1,2 and 3 correspond to 'avoid', 'inconsequential' and 'buy' respectively. The mean of each of these probabilities, $S(z_1)_{mean}, S(z_2)_{mean}, S(z_3)_{mean}$ is calculated for each asset on each unique day in testing. The class corresponding to the maximum of these mean probabilities is assigned to the entire day for each asset. If the maximum mean probability corresponds to the 'buy' class, then the asset is bought at the next-day open and sold at close as before.

Using this initial multi-class strategy, performance metrics of PP = 55.9\% and ATP = 0.240 \% are achieved. Hence by altering the task to multi-class and using a buying strategy as outlined, greater risk is observed than in the baseline \textit{multi-width} \textit{non-static} configuration. Hence, it is necessary to implement a buy threshold for the multi-class system. Instead of simply making a next-day investment if the maximum mean probability corresponds to the 'buy' class, implementation of a buy threshold also requires the mean probability $S(z_3)_{mean}$ to be greater than some buy threshold $t$. Therefore an investment is made if and only if the following requirements are met:

\begin{equation}
\max\limits_{k\in[1,3]}(S(z_k)_{mean}) = S(z_3)_{mean} > t
\end{equation}

\begin{figure}
\centering
\includegraphics[scale=0.8]{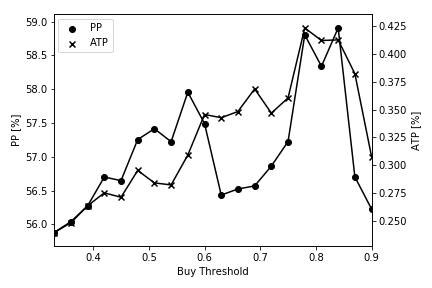}
\caption{Variation in PP and ATP with buy threshold across the entire date testing date range using a strategy developed from multi-class labelling. }
\label{multi_pp}
\end{figure}

Figure \ref{multi_pp} shows the variation in PP and ATP for the multi-class system with the implementation of buy thresholds $t \in [0.33, 0.9]$. The addition of a strict buy threshold results in a significant improvement in both metrics considered. Minimum risk is achieved using a buy threshold $t=0.86$ where although PP remains less than in the baseline \textit{multi-width} \textit{non-static} configuration, ATP is more than doubled. This suggests that although more incorrect trades are being executed, the average loss of the incorrect trades is reduced. The maximum single-day loss incurred using this method is 7.6\% compared to 11.3\% in the baseline and binary buy threshold strategies. Furthermore, the average return of correct buys is greater than in the baseline case due to the restriction that returns must be greater than 0.5\% for 'buy' allocation in multi-class labelling. The average return of correct buys using $t=0.86$ is 1.53\% compared to 1.20 \% in the baseline case. In summary, the implementation of a strict buy threshold with multi-class labelling decreases trading risk by minimising the average loss of incorrect next-day predictions, while the restriction on percentage returns in the labelling of individual headlines as 'buy' leads to the average profit of correct investments being far greater.


\section{Conclusion}
In this work, a Convolutional Neural Network is implemented to predict next-day stock fluctuations for three technology-based assets. Word-embeddings in three states; \textit{self-learnt}, \textit{static} and \textit{non-static}, are considered as well as \textit{single-width} and \textit{multi-width} convolutional layers. Experiments seeking to identify an optimal configuration in terms of accuracy and F1-score showed the presence of a filter width $h=3$ or $h=4$ as optimal. This result arises as key phrases depicting the headline's overall sentiment can be evaluated in their entirety. Word-embeddings in the \textit{non-static} state were found to be able to adapt to the specific context of the task whilst retaining relationships based on general context and hence provide the best classification performance. A \textit{multi-width} \textit{non-static} implementation was found to be the optimal configuration of the CNN architecture, leading to a testing accuracy of 61.7\%. However the benefit of using multiple filter widths compared to a single width was not compelling in the conducted experiments.

A simple trading strategy using the mean sigmoid prediction for headlines relating to each asset on each testing day was implemented. The optimal model configuration for classification was found to produce the best simulated trading results in terms of returns and two common trading performance metrics; PP and ATP. This configuration was found to more than triple an initial investment over the 838 testing day period.

Two methods to reduce the perceived risk in investments made across the testing set were developed. The implementation of a moderately strict buy threshold led to some reduction in risk, however further increase in this threshold resulted in increased risk compared to the baseline strategy. Alteration of the task to multi-class showed no reduction in risk on its own, but the combination of this with a strict buy threshold yielded an ATP more than double that achieved using the baseline strategy. Both of the discussed methods revealed downfalls of basing market predictions solely on company headlines. Therefore, combining methods presented in this work with predictions based on the technical analysis of stock trends should be considered in further research. Furthermore, headlines describing the general state of the economy could be considered in parallel to company-specific headlines. 

The training of the discussed model on separate collections of headlines grouped by business sector (oil and gas, finance etc.) should be undertaken in further work to form collections of embeddings and weights tuned to each sector. Subsequent testing headlines can then be evaluated using the set of weights and embeddings corresponding to the sector of the asset in question. This method has the ability to detect phrases based on the specific context of each sector as has been demonstrated for the technology sector in this work. However further work is needed to validate if this is beneficial compared to training a single collection of embeddings and weights for all sectors.

\vspace{1cm}
Conflict of Interest: The authors declare that they have no conflict of interest.

\bibliography{bibliography}{}
\bibliographystyle{plain}

\end{document}